\documentclass[10pt]{article} 
\usepackage[preprint]{tmlr}

\usepackage{amsmath,amsfonts,bm}

\def\eqref#1{equation~\ref{#1}}

\def\1{\bm{1}}

\DeclareMathAlphabet{\mathsfit}{\encodingdefault}{\sfdefault}{m}{sl}
\SetMathAlphabet{\mathsfit}{bold}{\encodingdefault}{\sfdefault}{bx}{n}

\usepackage{graphicx}
\usepackage{amsmath}
\usepackage{amssymb}
\usepackage{booktabs}
\usepackage{comment}
\usepackage{makecell}
\usepackage{amsfonts}
\usepackage{float}
\usepackage{multirow}
\usepackage{subcaption}
\usepackage{lipsum}
\usepackage{color, colortbl}
\usepackage{hyperref}
\usepackage{url}

\title{DiNAT-IR: Exploring Dilated Neighborhood Attention for High-Quality Image Restoration}

\author{\name Hanzhou Liu \email hanzhou1996@tamu.edu \\
      Texas A\&M University
      \AND
      \name Binghan Li \email lbh1994usa@tamu.edu \\
      Texas A\&M University
      \AND
      \name Chengkai Liu \email liuchengkai@tamu.edu\\
      Texas A\&M University
      \AND
      \name Mi Lu \email mlu@ece.tamu.edu  \\
      Texas A\&M University}

\newcommand\ith{\mathop{\mbox{$i$-}}}
\newcommand\bftab{\fontseries{b}\selectfont}

\newcommand{\ul}[1]{\underline{#1}}

\def\month{MM}  
\def\year{YYYY} 
\def\openreview{\url{https://openreview.net/forum?id=XXXX}} 

\begin{document}

\maketitle

\begin{abstract}
Transformers, with their self-attention mechanisms for modeling long-range dependencies, have become a dominant paradigm in image restoration tasks. However, the high computational cost of self-attention limits scalability to high-resolution images, making efficiency-quality trade-offs a key research focus. To address this, Restormer employs channel-wise self-attention, which computes attention across channels instead of spatial dimensions. While effective, this approach may overlook localized artifacts that are crucial for high-quality image restoration. 
\
To bridge this gap, we explore Dilated Neighborhood Attention (DiNA) as a promising alternative, inspired by its success in high-level vision tasks. DiNA balances global context and local precision by integrating sliding-window attention with mixed dilation factors, effectively expanding the receptive field without excessive overhead. However, our preliminary experiments indicate that directly applying this global-local design to the classic deblurring task hinders accurate visual restoration, primarily due to the constrained global context understanding within local attention. 
\
To address this, we introduce a channel-aware module that complements local attention, effectively integrating global context without sacrificing pixel-level precision. The proposed DiNAT-IR, a Transformer-based architecture specifically designed for image restoration, achieves competitive results across multiple benchmarks, offering a high-quality solution for diverse low-level computer vision problems.
\
Our codes have been released at \href{https://github.com/HanzhouLiu/DiNAT-IR}{https://github.com/HanzhouLiu/DiNAT-IR}.

\end{abstract}

\section{Introduction}

Image restoration is a fundamental task in computer vision, with wide-ranging applications in fields such as autonomous driving, medical imaging, and satellite remote sensing~\citep{ding2021perceptual, zhang2020review, rasti2021image}. It aims to recover high-quality images from degraded inputs, addressing challenges like blur, noise, and the other types of artifacts~\citep{banham1997digital}. 

In recent years, Transformers have emerged as powerful models for image restoration. Unlike traditional convolutional neural networks (CNNs) that rely on staked convolutional layers~\citep{zhang2017learning, zamir2021multi, chen2022simple}, Transformers utilize self-attention to model long-range pixel relationships~\citep{liang2021swinir, wang2022uformer, zamir2022restormer}, making them particularly effective for low-level computer vision tasks like deblurring, denoising, deraining, and super-resolution.

\begin{figure*}[tb]
  \centering
  \includegraphics[width=0.98\linewidth, trim = 0.5cm 13.8cm 6.05cm 0.52cm]{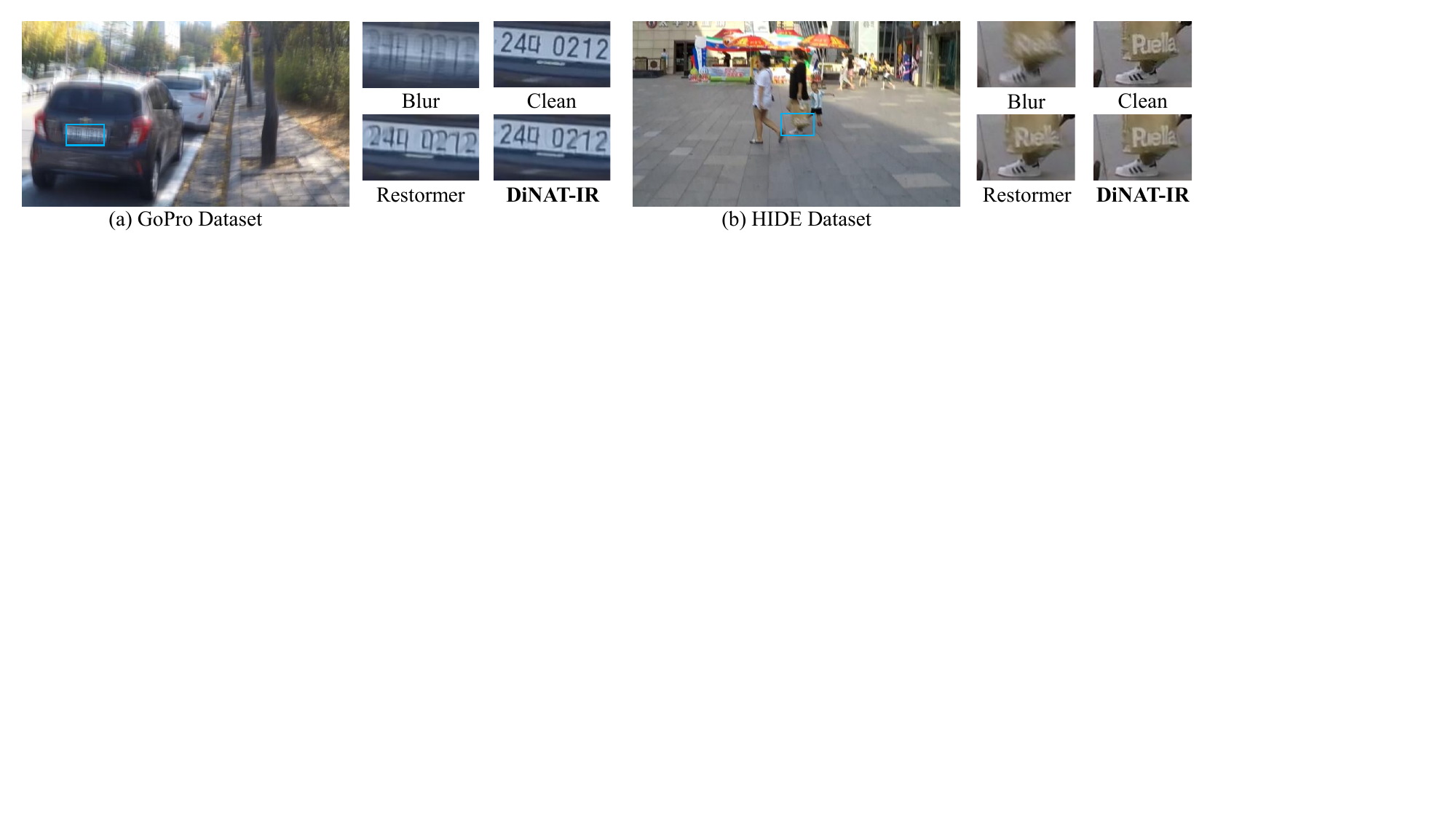}

  \caption{Visual comparisons between Restormer~\citep{zamir2022restormer} and our proposed DiNAT-IR on the motion deblurring datasets~\citep{nah2017deep, shen2019human}. DiNAT-IR produces cleaner restoration of numbers and characters on car license plates and hand-held bags. Zoom in to see details.}\label{fig:intro}
\end{figure*}

Despite their effectiveness, balancing the computational cost of self-attention with restoration quality remains a key challenge, especially for high-resolution images. Restormer~\citep{zamir2022restormer} addresses this by computing self-attention along the channel dimension instead of the spatial domain, achieving a strong trade-off between efficiency and performance. However, recent studies report that this design misses local details, as shown in Figure~\ref{fig:intro}, which are critical in dynamic scenes~\citep{jang2023spach, chen2024lgit}.

To bridge this gap, we explore Dilated Neighborhood Attention (DiNA) as a promising alternative, inspired by its recent success in detection and segmentation~\citep{hassani2022dilated}. Unlike previous self-attention mechanisms, which either aggregate global context entirely or focus solely on local patches, DiNA integrates sliding-window attention with mixed dilation factors, effectively expanding the receptive field without incurring excessive computational overhead.  The original DiNAT~\citep{hua2019dilated} reports that a hybrid design, using local neighborhood attention (NA) with a dilation factor $\delta=1$ alongside global DiNA, improves performance in high-level computer vision tasks.. However, our preliminary experiments reveal that directly applying this hybrid design to motion deblurring results in a notable performance drop compared to global-DiNA-only methods. We attribute this to the limited global context understanding of local NA, which restricts its ability to recover clean structures in full-resolution images.

To address this challenge, we introduce a channel-aware module that complements local attention by efficiently integrating global context without sacrificing pixel-level precision. This design effectively addresses the aforementioned bottleneck, allowing for more comprehensive feature interactions across the entire image. Furthermore, the proposed architecture, DiNAT-IR, has achieved competitive results across multiple benchmarks, demonstrating its potential as a high-fidelity solution for diverse image restoration challenges.

Our main contributions are threefold:
\begin{itemize}
    \item We investigate the application of dilated neighborhood attention for image deblurring and identify key limitations of its hybrid attention design in this context.
    \item We introduce a simple while effective channel-aware module that complements local neighborhood attention and restores global context without sacrificing pixel-level detail.
    \item We propose DiNAT-IR, a Transformer-based architecture that achieves competitive performance not only on deblurring benchmarks but also on other restoration tasks.
\end{itemize}
\section{Related Work}


\noindent \textbf{CNNs for Image Restoration.}
Convolutional neural networks (CNNs) consistently demonstrate strong performance across low-level computer vision tasks. DnCNN~\citep{zhang2017learning} pioneers the use of residual learning for image denoising, laying the groundwork for deeper and more effective architectures. MPRNet~\citep{zamir2021multi} adopts a multi-stage framework that processes image features at multiple spatial scales, achieving state-of-the-art results in image restoration. In the era of Transformer-based models, NAFNet~\citep{chen2022simple} stands out by showing that, with proper optimization, compact and purely convolutional architectures can still rival more complex Transformer designs in both efficiency and performance. Nevertheless, a key limitation of CNN-based approaches lies in their reliance on deeply stacked convolutional layers to enlarge the receptive field, which restricts their ability to model long-range dependencies effectively.

\noindent \textbf{Transformers for Image Restoration.}
In contrast, Transformer-based architectures inherently model global context through self-attention mechanisms. While applying vanilla Transformers~\citep{vaswani2017attention} to high-resolution images faced challenges due to the quadratic computational complexity of self-attention with respect to spatial dimensions, subsequent architectural innovations have significantly mitigated this issue in low-level computer vision tasks. For example, SwinIR~\citep{liang2021swinir} combines convolutional layers for shallow feature extraction with shifted window-based Transformer blocks to capture deeper representations, achieving strong performance in tasks such as super-resolution and denoising. Uformer~\citep{wang2022uformer} integrates Locally-enhanced Window (LeWin) attention within a U-Net structure, effectively preserving spatial detail for deblurring and deraining tasks. Different from window-based methods, Restormer~\citep{zamir2022restormer} improves computational efficiency by computing self-attention along the channel dimension rather than spatial dimensions. However, follow-up studies~\citep{jang2023spach, chen2024lgit} observe that such designs may overlook fine-grained local details that are critical for restoration in real-world environments.

\noindent \textbf{Dilated Neighborhood Attention.}
Recent advancements in vision Transformers have prioritized improving the efficiency of self-attention mechanisms while preserving their ability to capture long-range dependencies. Hierarchical models such as the Swin Transformer~\citep{liu2021swin} and the Neighborhood Attention Transformer~\citep{hassani2023neighborhood} reduce computational costs by restricting self-attention to local windows. However, this often comes at the expense of the global receptive field, an essential attribute for high-level visual understanding. To address this limitation, \textit{Dilated Neighborhood Attention} ({\bftab DiNA})~\citep{hassani2022dilated} extends \textit{neighborhood attention} ({\bftab NA}) by sparsifying it across dilated local regions. This design enables an exponential increase in the receptive field without incurring additional computational overhead. Their resulting model, the Dilated Neighborhood Attention Transformer, with dense local {\bftab NA}  and sparse global {\bftab DiNA} (abbreviated as {\bftab NA-DiNA}), achieving strong performance in high-level computer vision tasks such as object detection, instance segmentation, and semantic segmentation.
\
Despite these strengths, we observe that directly applying the original {\bftab NA-DiNA} method to low-level computer vision tasks like motion deblurring results in a noticeable performance drop compared to the {\bftab DiNA-only} attention design. This may be attributed to the inherently limited global context understanding of local {\bftab NA}, which struggles to fully capture the spatial extent and complexity of image degradation patterns common in motion blur scenarios.

\section{Method}
The overall pipeline of DiNAT-IR is based on Restormer~\citep{zamir2022restormer}. It adopts a multi-level U-Net structure that efficiently captures degradation patterns through hierarchical feature processing. The encoder gradually downsamples the input to extract deep features, while the decoder upsamples and refines the output using skip connections that preserve spatial resolution. We build upon this framework and integrate an improved attention mechanism, which is detailed in the following sections.

\definecolor{darkblue}{RGB}{46,117,182}
\definecolor{darkgreen}{RGB}{84,130,53}
\subsection{Alternating NA-DiNA Attention Scheme}\label{sec:ads}

\begin{figure*}[tb]
  \centering
  \includegraphics[width=0.98\linewidth, trim = 0.5cm 7.2cm 7.05cm 0.52cm]{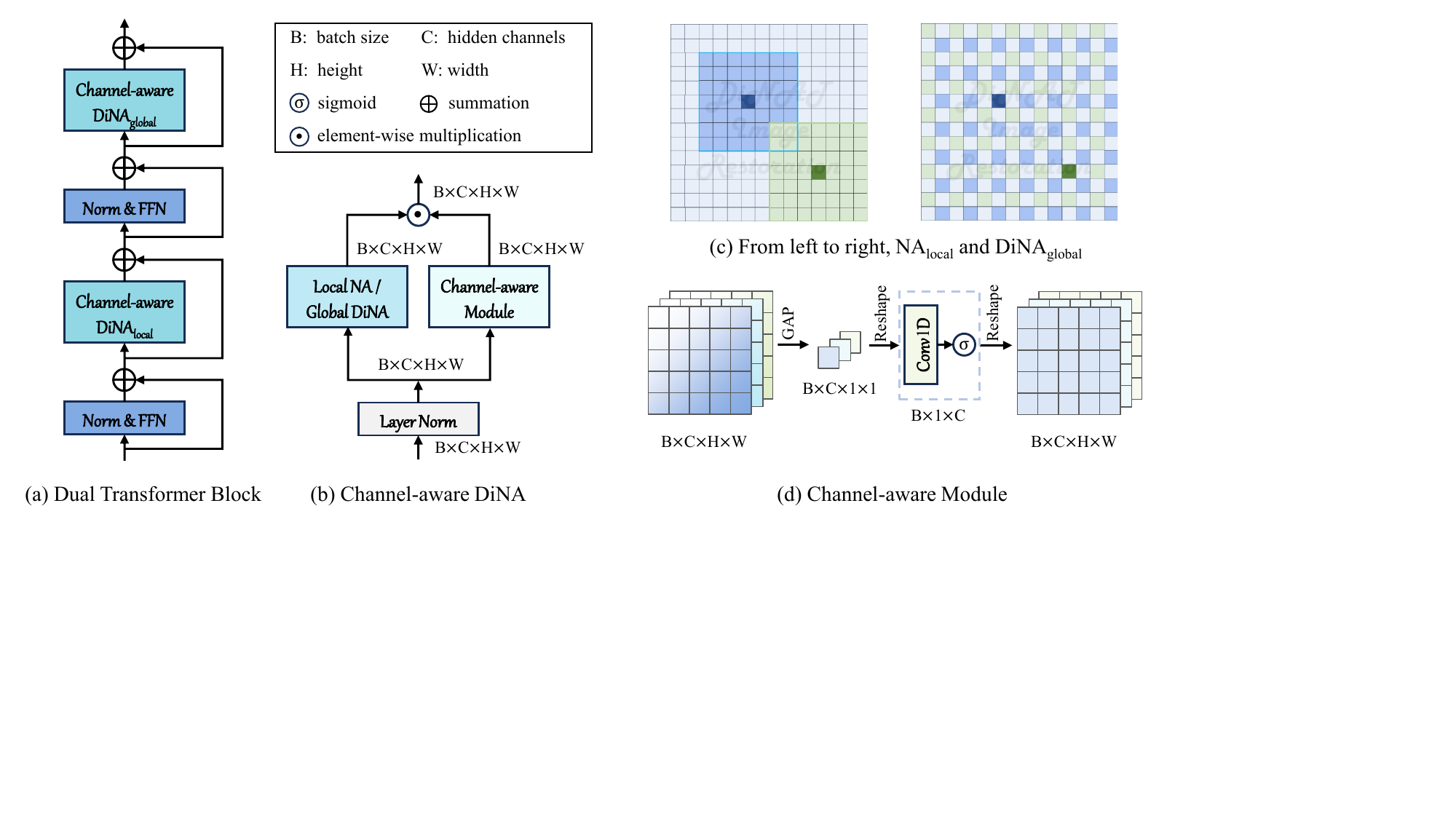}

    \caption{Structures of (a) the Dual Transformer block with the alternating NA-DiNA attention scheme, (b) the channel-aware DiNA module, (c) local DiNA (NA) and global DiNA (DiNA) blocks, and (d) the channel-aware module. Note: GAP denotes global average pooling, and Conv1D indicates 1D convolution.}\label{fig:method_blocks}
\end{figure*}

To effectively model both fine-grained structures and large-scale degradation patterns, DiNAT-IR integrates an alternating NA-DiNA strategy within its Transformer blocks, drawing inspiration from the Dilated Neighborhood Attention Transformer (DiNAT)~\citep{hassani2022dilated}. By setting the dilation factor $\delta$ to 1, DiNA effectively reduces to standard Neighborhood Attention (NA)~\citep{hassani2023neighborhood}. At each level of DiNAT-IR, the self-attention blocks alternate between two dilation factors to vary the attention window size. Specifically, setting the dilation factor $\delta=1$ yields local NA, while larger values of $\delta$ expand the receptive field to capture broader context. The dilation pairs are defined as $\delta \in \{1, 36\}$, $\{1, 18\}$, $\{1, 9\}$, and $\{1, 4\}$ across the four stages of the network, corresponding to progressively finer spatial resolutions. This alternating pattern allows DiNAT-IR to adaptively integrate both local details and global contextual information, improving its capacity to model spatially extensive degradations without introducing significant computational overhead.

While the original {\bftab NA-DiNA} architecture was developed for high-level vision tasks, its hybrid attention design can also be intuitively extended to image restoration problems. In this context, the local {\bftab NA} is expected to model short-range, pixel-level dependencies, while the sparse {\bftab DiNA} captures broader degradation patterns. However, our preliminary experiments reveal that directly applying the vanilla {\bftab NA-DiNA} configuration to the low-level tasks such as motion deblurring leads to a noticeable performance drop compared to a {\bftab DiNA-only} baseline. We attribute it to the significantly reduced global context understanding introduced by the frequent use of local {\bftab NA}. To address this limitation, we propose a lightweight channel-aware module designed to preserve global context modeling while mitigating the drawbacks of overly localized attention.

\subsection{Channel Aware Self Attention}
Figure~\ref{fig:method_blocks} (a) shows that channel-aware self-attention contains two parallel units, the self-attention layers (SA) and a channel-aware module (CAM). DiNAT-IR uses alternating neighborhood attention ({\bftab NA}) and dilated neighborhood attention ({\bftab DiNA}) as the basic component of SA. Furthermore, CAM is proposed to solve the issue of limited receptive filed caused by {\bftab NA}. As illustrated in Figure~\ref{fig:method_blocks} (c), a CAM first transforms the normalized 2-D features into 1-D data by global average pooling (GAP)~\citep{lin2013network, chu2022improving}; then, it applies a 1-D convolution to the intermediate features along the channel dimension; finally, a $\emph{sigmoid}$ function is adopted to compute attention scores. The outputs of the CAM and DiNA\@ are merged by element-wise multiplications.

Given a layer normalized input tensor $\mathbf{X}\in\mathbb{R}^{\hat{H}\times\hat{W}\times\hat{C}}$.
The output of CASA is computed as,
\begin{equation}
  \begin{aligned} \label{eq_CAM}
    \hat{\mathbf{X}} &= \mathrm{SA}(\mathbf{X})\odot \mathrm{CAM}(\mathbf{X}),\\
  {\rm CAM}(\mathbf{X}) &= f^{-1}(\mathrm{Conv}_{1d}(f(\mathrm{GAP}_{2d}^{1\times1}(\mathbf{X})))),
  \end{aligned}
\end{equation}
where $\odot$ denotes element-wise multiplication; $f$ is a tensor manipulation function which squeezes and transposes a $C\times 1\times 1$ matrix, resulting in a $1\times C$ matrix; $\mathrm{Conv}_{1d}$ denotes a 1D convolution with a kernel size of $3$; ${\rm GAP}_{2d}^{1\times1}$ indicates global average pooling, outputting a tensor of size $1\times1$. We employed the GAP design proposed by~\citeauthor{chu2022improving}, and the CAM idea draws inspiration from ECA-Net~\citep{wang2020eca}.


\section{Experiments and Analysis}
We evaluate the performance of DiNAT-IR across four distinct image restoration tasks: (a) single-image motion deblurring, (b) defocus deblurring with dual inputs and single images, (c) single image deraining, and (d) single image denoising. In the result tables, the best-performing and second-best methods are indicated using {\bftab bold} and \ul{underline} formatting respectively. We primarily compare against multi-task image restoration networks, supplemented by task-specific methods for completeness.

\noindent
{\bftab Implementation Details.} DiNAT-IR adopts the four-stage U-Net architecture of Restormer~\citep{zamir2022restormer} as its backbone. All experiments are conducted using a batch size of 16 across 8 NVIDIA A100 GPUs. Task-specific training configurations vary depending on the particular restoration task and dataset.

\noindent
{\bftab GoPro}~\citep{nah2017deep}. For the motion deblurring task, we train DiNAT-IR with image patches of size $256 \times 256$ and a batch size of 16 for 600K iterations using PSNR loss. The initial learning rate is set to $3 \times 10^{-4}$ and gradually reduced to $1 \times 10^{-6}$ following a cosine annealing schedule. We use AdamW as the optimizer with betas set to $[0.9, 0.999]$~\citep{loshchilov2017decoupled}. We further fine-tune the network with an image size of $384 \times 384$ and a batch size of 8 for an additional 200K iterations, inspired by the progressive training strategy employed in Restormer~\citep{zamir2022restormer} and Stripformer~\citep{tsai2022stripformer}. During fine-tuning, the initial learning rate is set to $1 \times 10^{-4}$. We observe that DiNAT-IR may not fully converge to an optimal solution, suggesting that improved training strategies could further enhance performance on the GoPro dataset. The final model used for evaluation is obtained from the last training iteration.

\noindent
{\bftab DPDD}~\citep{abuolaim2020defocus}. For the dual-pixel defocus deblurring task, the dual-input variant of DiNAT-IR is trained with an image size of $256 \times 256$ and a batch size of 16 for 300K iterations using PSNR loss. The optimizer and learning rate schedule are consistent with those used for motion deblurring. The model at the 290K iteration is selected as our dual-pixel defocus deblurring model. For the single-image defocus deblurring task, we re-train DiNAT-IR with single images as inputs and adopt the model checkpoint at the 140K iteration as the final version.

\noindent
{\bftab Rain13K}~\citep{jiang2020multi}. For the single image deraining task, DiNAT-IR is trained with an image size of $256 \times 256$ and a batch size of 16 for 300K iterations using L1 loss. The optimizer and learning rate schedule follow the same settings as in motion deblurring. We further fine-tune the network with an image size of $384 \times 384$ and a batch size of 8 for an additional 100K iterations, selecting the model at the 40K fine-tuning iteration for our deraining experiments.

\noindent
{\bftab SIDD}~\citep{abdelhamed2018high}. For the real-world image denoising task, DiNAT-IR is trained with an image size of $256 \times 256$ and a batch size of 16 for 300K iterations using PSNR loss. The optimizer and learning rate schedule are identical to those in the motion deblurring task. We choose the model at the 220K iteration as our final denoising model.

\setlength{\tabcolsep}{8pt}
\begin{table}[ht]
\centering
\caption{Comparisons of image restoration models on GoPro~\citep{nah2017deep} and HIDE~\citep{shen2019human} datasets. We follow MaIR~\citep{MaIR} and report PSNR, SSIM, Params (M), and FLOPs (G).The proposed DiNAT-IR has achieved competitive performance compared to recent restoration networks.}
\begin{tabular}{lcccccc}
\toprule
\multirow{2}{*}{Method} & \multicolumn{2}{c}{GoPro} & \multicolumn{2}{c}{HIDE} & \multicolumn{2}{c}{Model Complexity} \\
\cmidrule(lr){2-3} \cmidrule(lr){4-5} \cmidrule(lr){6-7}
& PSNR ↑ & SSIM ↑ & PSNR ↑ & SSIM ↑ & Params (M) ↓ & FLOPs (G) ↓ \\
\midrule
\rowcolor{gray!20}
SRN~\textcolor{red}{\citeyear{tao2018scale}}    
& 30.26 & 0.934 & 28.36 & 0.904 & \ul{3.76}   & 35.87 \\
DBGAN~\textcolor{red}{\citeyear{zhang2020deblurring}}          
& 31.10 & -     & 28.94 & -     & 11.59  & 379.92 \\
\rowcolor{gray!20}
MT-RNN~\textcolor{red}{\citeyear{park2020multi}}     
& 31.15 & -     & 29.15 & -     & \bftab2.64   & \bftab13.72 \\
DMPHN~\textcolor{red}{\citeyear{zhang2022accurate}}       
& 31.20 & -     & 29.09 & -     & 86.80  & - \\
\rowcolor{gray!20}
CODE~\textcolor{red}{\citeyear{zhao2023comprehensive}}        
& 31.94 & -     & 29.67 & -     & 12.18  & 22.52 \\
MIMO~\textcolor{red}{\citeyear{cho2021rethinking}}
& 32.45 & 0.956 & 29.99 & 0.930 & 16.10  & 38.64 \\
\rowcolor{gray!20}
MPRNet~\textcolor{red}{\citeyear{zamir2021multi}}
& 32.66 & 0.958 & 30.96 & 0.939 & 20.13  & 194.42 \\
Restormer~\textcolor{red}{\citeyear{zamir2022restormer}}  
& 32.92 & 0.961 & 31.22 & 0.942 & 26.13  & 35.31 \\
\rowcolor{gray!20}
Uformer~\textcolor{red}{\citeyear{wang2022uformer}}
& 33.06 & 0.967 & 30.90 & \bftab 0.953 & 50.88  & 22.36 \\
CU-Mamba~\textcolor{red}{\citeyear{deng2024cu}}
& 33.53 & -     & \ul{31.47} & -     & 19.70  & - \\
\rowcolor{gray!20}
NAFNet~\textcolor{red}{{\citeyear{chen2022simple}}}
& 33.69 & 0.966 & 31.32 & 0.942 & 67.89  & \ul{15.85} \\
MaIR~\textcolor{red}{\citeyear{MaIR}}  
& \ul{33.69} & \bftab0.969 & 31.57 & \ul{0.946} & 26.29  & 49.29 \\ 
\midrule
\rowcolor{gray!20}
\bftab DiNAT-IR 
& \bftab33.80 & \ul{0.967} & \bftab31.57 & 0.945 & 25.90  & 45.62 \\
\bottomrule
\end{tabular}
\label{tab:model_gopro_hide}
\end{table}

\subsection{Motion Deblurring Results}
\begin{figure*}[tb]
  \centering
  \includegraphics[width=0.98\linewidth, trim = 0.cm 14.8cm 12.6cm 0.cm]{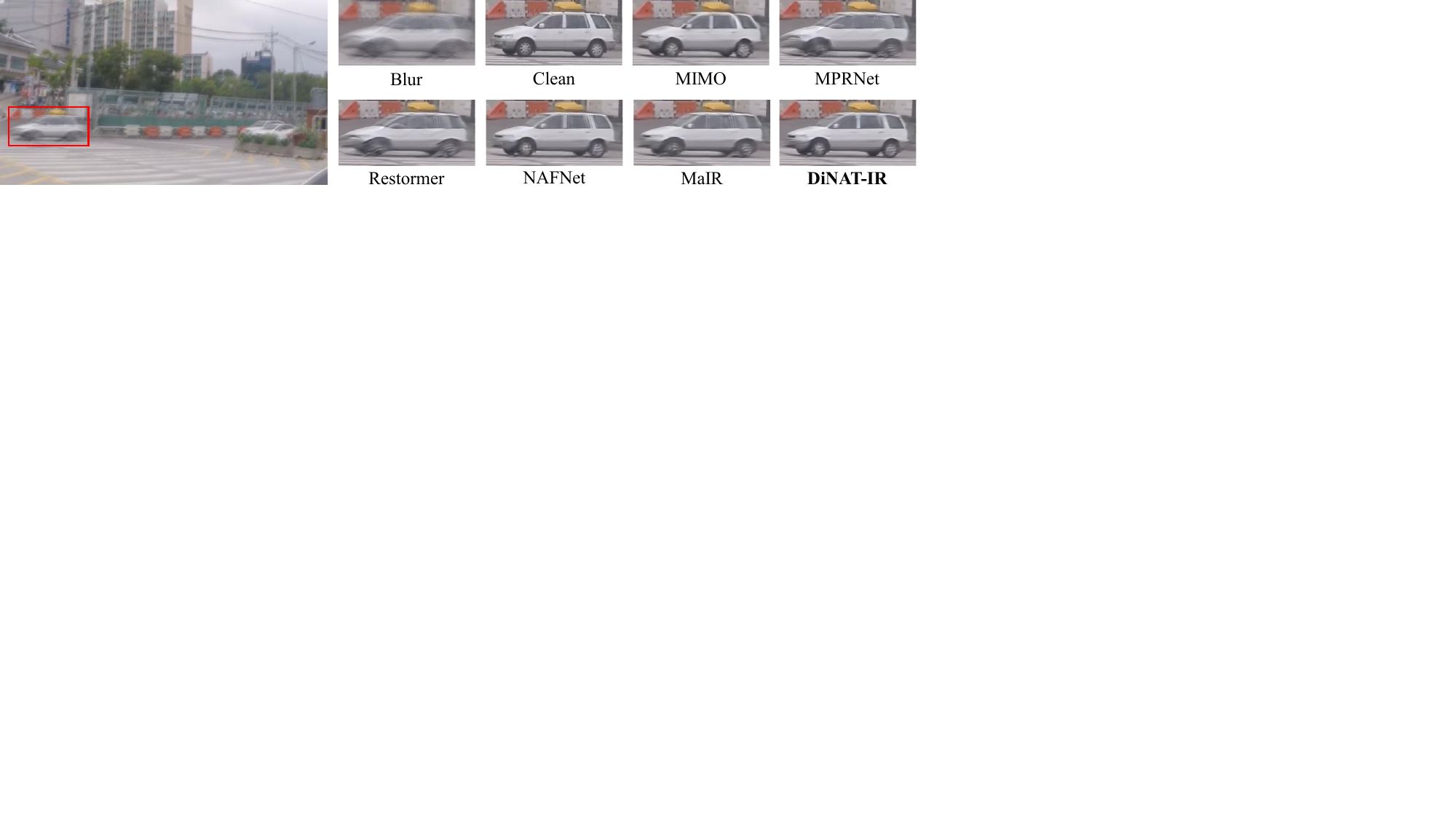}

  \caption{Deblurring results on the GoPro dataset~\citep{nah2017deep}. Zoom in to see details.}\label{fig:sota_gopro}
\end{figure*}

\begin{figure*}[tb]
  \centering
  \includegraphics[width=0.98\linewidth, trim = 0.cm 14.6cm 2cm 0.cm]{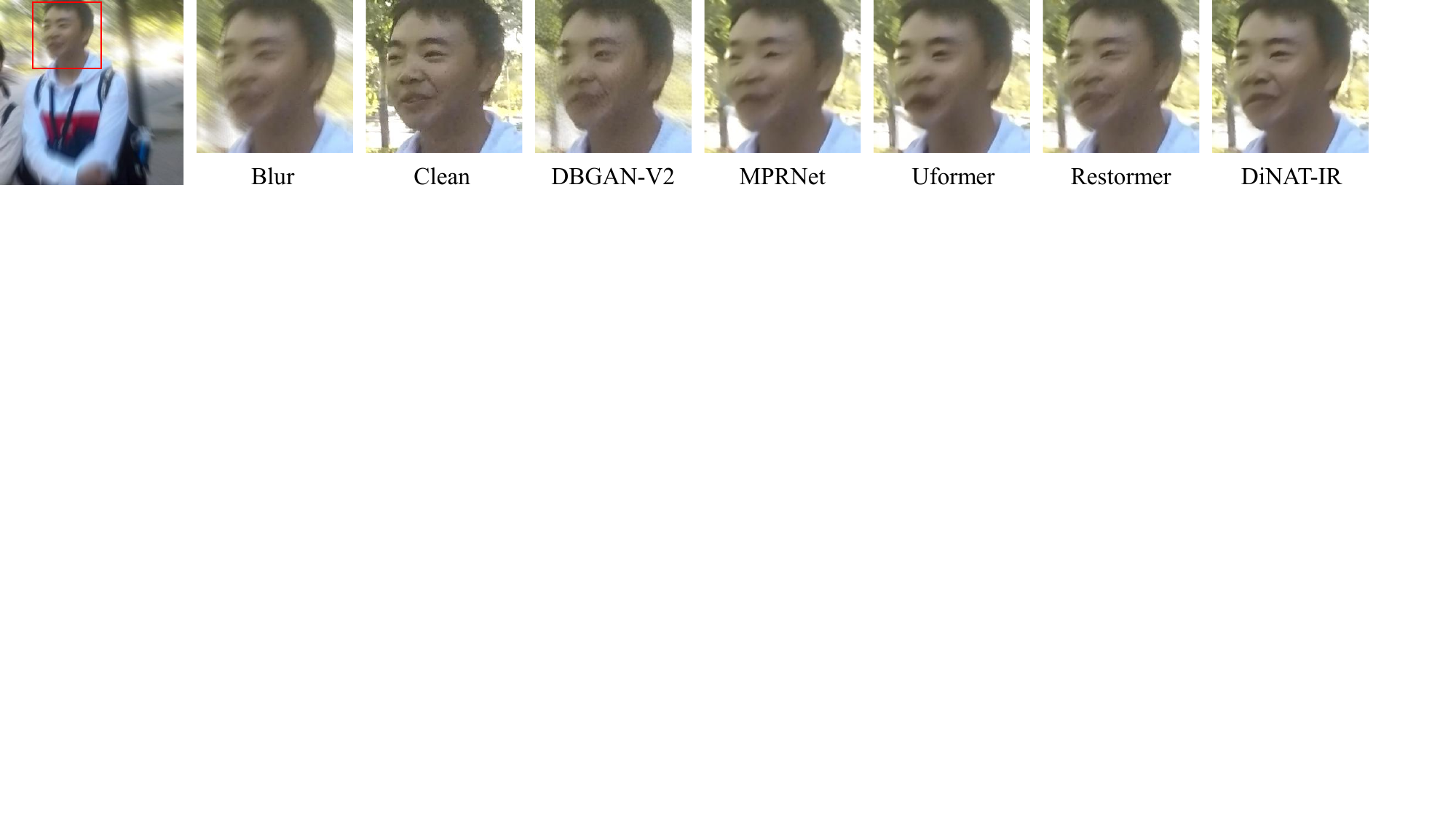}

  \caption{Deblurring results on the HIDE dataset~\citep{shen2019human}. Zoom in to see details.}\label{fig:sota_hide}
\end{figure*}

We conduct a thorough evaluation of various image restoration models on the GoPro~\citep{nah2017deep} and HIDE~\citep{shen2019human}) datasets. As summarized in Table~\ref{tab:model_gopro_hide}, our proposed DiNAT-IR consistently achieves strong results across the board. Specifically, it reaches a PSNR of 33.80 dB on GoPro and 31.57 dB on HIDE, matching or surpassing all compared methods, including the recent high-performing MaIR~\citep{MaIR} and NAFNet\citep{chen2022simple}. While MaIR reports a comparable PSNR on both datasets, DiNAT-IR achieves this with slightly fewer parameters and competitive FLOPs. Compared to traditional convolution-based models like MPRNet~\citep{zamir2021multi} and attention-based method Restormer~\citep{zamir2022restormer}, DiNAT-IR maintains similar or superior accuracy while preserving efficiency. Furthermore, despite being trained solely on GoPro, DiNAT-IR demonstrates excellent generalization to HIDE, underscoring its robustness in human-centric scenarios. These results highlight DiNAT-IR’s effective trade-off between model complexity and restoration quality, as well as its potential as a strong alternative in dynamic scenes.
Qualitative comparisons in Figure~\ref{fig:sota_gopro} and Figure~\ref{fig:sota_hide} clearly demonstrate that the image deblurred by our
method is more visually closer to the ground-truth than those of the other algorithms.

\setlength{\tabcolsep}{4.4pt}
\begin{table}[t]
\centering
\caption{{\bf Dual-Pixel Defocus Deblurring comparisons} on the DPDD dataset~\citep{abuolaim2020defocus}, which includes 37 indoor and 39 outdoor scenes. $D$ indicates network variants using dual-image inputs; $S$ denotes the single-image task. DiNAT-IR demonstrates performance comparable to GRL-B~\citep{li2023efficient} across both single-image and dual-pixel settings. }
\begin{tabular}{l ccc ccc ccc}
\toprule
\multirow{2}{*}{Method} & \multicolumn{3}{c}{Indoor Scenes} & \multicolumn{3}{c}{Outdoor Scenes} & \multicolumn{3}{c}{Combined} \\
\cmidrule(lr){2-4} \cmidrule(lr){5-7} \cmidrule(lr){8-10}
& PSNR ↑ & SSIM ↑ & MAE ↓ & PSNR ↑ & SSIM ↑ & MAE ↓ & PSNR ↑ & SSIM ↑ & MAE ↓ \\
\midrule
EBDB$_S$~\textcolor{red}{\citeyear{karaali2017edge}}
& 25.77 & 0.772 & 0.040 & 21.25 & 0.599 & 0.058 & 23.45 & 0.683 & 0.049 \\
\rowcolor{gray!20}
DMENet$_S$~\textcolor{red}{\citeyear{lee2019deep}}
& 25.50 & 0.788 & 0.038 & 21.43 & 0.644 & 0.063 & 23.41 & 0.714 & 0.051 \\
JNB$_S$~\textcolor{red}{\citeyear{shi2015just}}
& 26.73 & 0.828 & 0.031 & 21.10 & 0.608 & 0.064 & 23.84 & 0.715 & 0.048 \\
\rowcolor{gray!20}
DPDNet$_S$~\textcolor{red}{\citeyear{abuolaim2020defocus}}
& 26.54 & 0.816 & 0.031 & 22.25 & 0.682 & 0.056 & 24.34 & 0.747 & 0.044 \\
KPAC$_S$~\textcolor{red}{\citeyear{son2021single}}
& 27.97 & 0.852 & 0.026 & 22.62 & 0.701 & 0.053 & 25.22 & 0.774 & 0.040 \\
\rowcolor{gray!20}
IFAN$_S$~\textcolor{red}{\citeyear{lee2021iterative}} 
& 28.11 & 0.861 & 0.026 & 22.76 & 0.720 & 0.052 & 25.37 & 0.789 & 0.039 \\
Restormer$_S$~\textcolor{red}{\citeyear{zamir2022restormer}}
& 28.87 & 0.882 & 0.025 & 23.24 & 0.743 & 0.050 & 25.98 & 0.811 & 0.038 \\
\rowcolor{gray!20}
CSformer$_S$~\textcolor{red}{\citeyear{duan2023masked}}
& \ul{29.01} & \ul{0.883} & \bftab 0.023 & \bftab 23.63 & \ul{0.759} & \bftab 0.047 & \bftab 26.25 & \ul{0.819} & \bftab 0.036 \\
GRL-B$_S$~\textcolor{red}{\citeyear{li2023efficient}}
& \bftab 29.06 & \bftab 0.886 & \ul{0.024} & 23.45 & \bftab 0.761 & 0.049 & \ul{26.18} & \bftab 0.822 & 0.037 \\
\rowcolor{gray!20}
\bftab DiNAT-IR$_S$ 
& 28.94 & 0.881 & 0.025 & \ul{23.48} & 0.751 & \ul{0.049} & 26.14 & 0.814 & \ul{0.037} \\
\midrule
DPDNet$_D$~\textcolor{red}{\citeyear{abuolaim2020defocus}}
& 27.48 & 0.849 & 0.029 & 22.90 & 0.726 & 0.052 & 25.13 & 0.786 & 0.041 \\
\rowcolor{gray!20}
RDPD$_D$~\textcolor{red}{\citeyear{abuolaim2021learning}}
& 28.10 & 0.843 & 0.027 & 22.82 & 0.704 & 0.053 & 25.39 & 0.772 & 0.040 \\
Uformer$_D$~\textcolor{red}{\citeyear{wang2022uformer}}
& 28.23 & 0.860 & 0.026 & 23.10 & 0.728 & 0.051 & 25.65 & 0.795 & 0.039 \\
\rowcolor{gray!20}
IFAN$_D$~\textcolor{red}{\citeyear{lee2021iterative}}
& 28.66 & 0.868 & 0.025 & 23.46 & 0.743 & 0.049 & 25.99 & 0.804 & 0.037 \\
Restormer$_D$~\textcolor{red}{\citeyear{zamir2022restormer}}
& 29.48 & 0.895 & 0.023 & 23.97 & 0.773 & \ul{0.047} & 26.66 & 0.833 & \ul{0.035} \\
\rowcolor{gray!20}
CSformer$_D$~\textcolor{red}{\citeyear{duan2023masked}}
& 29.54 & 0.896 & 0.023 & 24.38 & \ul{0.788} & 0.045 & 26.89 & 0.841 & 0.034 \\
GRL-B$_D$~\textcolor{red}{\citeyear{li2023efficient}}
& \bftab 29.83 & \bftab 0.903 & \bftab 0.022 & \ul{24.39} & 0.795 & 0.045 & 27.04 & \bftab 0.847 & 0.034 \\
\rowcolor{gray!20}
\bftab DiNAT-IR$_D$ 
& \ul{29.76} & \ul{0.901} & \ul{0.023} & \bftab 24.47 & \bftab 0.795 & \bftab 0.045 & \bftab 27.05 & \ul{0.846} & \bftab 0.034 \\
\bottomrule
\end{tabular}
\label{tab:dpdd_deblurring}
\end{table}

\subsection{Defocus Deblurring Results}
Table~\ref{tab:dpdd_deblurring} presents a comprehensive comparison of single-image and dual-pixel defocus deblurring methods on the DPDD dataset~\citep{abuolaim2020defocus}. For single-image defocus deblurring, DiNAT-IR$_S$ delivers competitive results, achieving strong performance across all metrics. It obtains the second-highest PSNR and MAE on outdoor scenes and ranks closely behind GRL-B$_S$~\citep{li2023efficient} and CSformer$_S$~\citep{duan2023masked} overall. Notably, while GRL-B$_S$ slightly surpasses DiNAT-IR$_S$ in combined PSNR (26.18 dB vs. 26.14 dB), DiNAT-IR$_S$ demonstrates comparable or better PSNR and SSIM on the outdoor scene.

In the dual-pixel setting, DiNAT-IR$_D$ shows excellent performance, either outperforming or closely matching state-of-the-art methods. It achieves the highest PSNR on outdoor scenes (24.47 dB) and the best combined PSNR (27.05 dB), while maintaining competitive SSIM and lowest MAE scores. Compared to Restormer$_D$, which performs strongly indoors, DiNAT-IR$_D$ offers superior outdoor performance and better balance across scenes. These results highlight DiNAT-IR’s capability to handle both single-image and dual-pixel defocus deblurring tasks effectively, achieving state-of-the-art performance on the DPDD benchmark.

\setlength{\tabcolsep}{2.4pt}
\begin{table}[t]
\centering
\caption{{\bf Image deraining results.} DiNAT-IR achieves performance very close to that of Restormer~\citep{zamir2022restormer}, with SSIM scores nearly matching those of Restormer across multiple datasets. However, we acknowledge noticeably lower PSNR scores for DiNAT-IR on these datasets.}
\begin{tabular}{l ccc ccc ccc ccc ccc}
\toprule
\multirow{2}{*}{Method} 
& \multicolumn{2}{c}{Rain100H} 
& \multicolumn{2}{c}{Rain100L} 
& \multicolumn{2}{c}{Test2800} 
& \multicolumn{2}{c}{Test1200} 
& \multicolumn{2}{c}{Test100} \\
\cmidrule(lr){2-3}
\cmidrule(lr){4-5}
\cmidrule(lr){6-7}
\cmidrule(lr){8-9}
\cmidrule(lr){10-11}
& PSNR ↑ & SSIM ↑ 
& PSNR ↑ & SSIM ↑ 
& PSNR ↑ & SSIM ↑ 
& PSNR ↑ & SSIM ↑ 
& PSNR ↑ & SSIM ↑ \\
\midrule
SEMI~\textcolor{red}{\citeyear{wei2019semi}}
& 16.56 & 0.486 & 25.03 & 0.842 & 24.43 & 0.782 & 26.05 & 0.822 & 22.35 & 0.788 \\
\rowcolor{gray!20}
DIDMDN~\textcolor{red}{\citeyear{zhang2018density}}
& 17.35 & 0.524 & 25.23 & 0.741 & 28.13 & 0.867 & 29.65 & 0.901 & 22.56 & 0.818 \\
UMRL~\textcolor{red}{\citeyear{yasarla2019uncertainty}}
& 26.01 & 0.832 & 29.18 & 0.923 & 29.97 & 0.905 & 30.55 & 0.910 & 24.41 & 0.829 \\
\rowcolor{gray!20}
RESCAN~\textcolor{red}{\citeyear{li2018recurrent}}
& 26.36 & 0.786 & 29.80 & 0.881 & 31.29 & 0.904 & 30.51 & 0.882 & 25.00 & 0.835 \\
PreNet~\textcolor{red}{\citeyear{ren2019progressive}}
& 26.77 & 0.858 & 32.44 & 0.950 & 31.75 & 0.916 & 31.36 & 0.911 & 24.81 & 0.851 \\
\rowcolor{gray!20}
MSPFN~\textcolor{red}{\citeyear{jiang2020multi}}
& 28.66 & 0.860 & 32.40 & 0.933 & 32.82 & 0.930 & 32.39 & 0.916 & 27.50 & 0.876 \\
MPRNet~\textcolor{red}{\citeyear{zamir2021multi}}
& 30.41 & 0.890 & 36.40 & 0.965 & 33.64 & 0.938 & 32.91 & 0.916 & 30.27 & 0.897 \\
\rowcolor{gray!20}
SPAIR~\textcolor{red}{\citeyear{purohit2021spatially}}
& 30.95 & 0.892 & 36.93 & 0.969 & 33.34 & 0.936 & \ul{33.04} & 0.922 & 30.35 & 0.909 \\
{\bf Restormer}~\textcolor{red}{\citeyear{zamir2022restormer}}
             & \bftab 31.46 & \bftab 0.904 
             & \bftab 38.99 & \bftab 0.978 
             & \bftab 34.18 & \bftab 0.944 
             & \bftab 33.19 & \bftab 0.926 
             & \bftab 32.00 & \bftab 0.923 \\
\rowcolor{gray!20}
{\bf DiNAT-IR}  
             & \ul{31.26} & \ul{0.903} 
             & \ul{38.93} & \ul{0.977} 
             & \ul{33.91} & \ul{0.943}
             & 32.31 & \ul{0.923}
             & \ul{31.22} & \ul{0.920} \\
\bottomrule
\end{tabular}
\label{tab:deraining_results}
\end{table}

\subsection{Deraining Results}
Table~\ref{tab:deraining_results} summarizes the performance of several image deraining models across five benchmark datasets. DiNAT-IR demonstrates excellent results, achieving SSIM scores nearly identical to those of Restormer~\citep{zamir2022restormer} across all five datasets, indicating strong perceptual quality and effective detail preservation. Although DiNAT-IR’s PSNR is slightly lower than Restormer’s, the differences are minor, for example, on the Rain100L test set, DiNAT-IR attains 38.93 dB compared to Restormer’s 38.99 dB, a negligible gap considering the task complexity. Compared to earlier methods such as SEMI~\citep{wei2019semi}, DIDMDN~\citep{zhang2018density}, and UMRL~\citep{yasarla2019uncertainty}, DiNAT-IR delivers significant improvements in both PSNR and SSIM. It also performs competitively against recent models like MPRNet~\citep{zamir2021multi} and SPAIR~\citep{purohit2021spatially}, surpassing them in several metrics. Overall, these results highlight DiNAT-IR as a highly effective deraining model, delivering competitive perceptual quality and achieving performance close to that of Restormer in pixel-level restoration accuracy.

\begin{table}[t]
\centering
\caption{{\bftab Real image denoising results.} All methods are trained and tested on the SIDD dataset~\citep{abdelhamed2018high}. DiNAT-IR achieves performance comparable to Restormer~\citep{zamir2022restormer}.}
\label{tab:sidd_comparison}
\begin{tabular}{lcccccccccc}
\toprule
\multirow{2}{*}{\textbf{Method}} 
& DnCNN & BM3D & VDN & MIRNet & MPRNet & DAGL & Uformer & Restormer & MambaIR & \bftab DiNAT-IR \\
&\citeyear{zhang2017beyond} &\citeyear{dabov2007image} &\citeyear{yue2019variational} & \citeyear{zamir2020learning} &\citeyear{zamir2021multi} &\citeyear{mou2021dynamic} &\citeyear{wang2022uformer} &\citeyear{zamir2022restormer} &\citeyear{guo2025mambair} & (Ours)\\
\midrule
\bftab PSNR ↑
& 23.66 & 25.65 & 39.28 & 39.72 & 39.71 & 38.94 & 39.89 & {\bftab 40.02} & 39.89 & \ul{39.89}\\
\bftab SSIM ↑
& 0.583 & 0.685 & 0.956 & 0.959 & 0.958 & 0.953 &  0.960 & 0.960 & 0.960 & {\bftab 0.960}\\
\bottomrule
\end{tabular}
\end{table}

\subsection{Denoising Results}
Table~\ref{tab:sidd_comparison} compares several real-image denoising methods based on PSNR and SSIM metrics. Early approaches like DnCNN~\citep{zhang2017beyond} and BM3D~\citep{dabov2007image} achieve substantially lower performance, with PSNRs below 26 dB and SSIM under 0.70, reflecting limited effectiveness on challenging real-world noise. Modern deep networks such as VDN~\citep{yue2019variational}, MIRNet~\citep{zamir2020learning}, MPRNet~\citep{zamir2021multi}, DAGL~\citep{mou2021dynamic}, and Uformer~\citep{wang2022uformer} demonstrate significant improvements, achieving PSNR values around $39\text{--}40$ dB and SSIM above 0.95, highlighting the advances brought by learning-based architectures. Among these methods, Restormer~\citep{zamir2022restormer} achieves the highest performance, with a PSNR of 40.02 dB and an SSIM of 0.960, establishing a strong benchmark. MambaIR~\citep{guo2025mambair} achieves a PSNR of 39.89 dB and an SSIM of 0.960, closely following Restormer. Our method, DiNAT-IR, also achieves highly competitive results, matching the highest SSIM score of 0.960 and attaining a PSNR of 39.89 dB. Overall, DiNAT-IR performs at the same level as MambaIR, demonstrating strong capabilities in preserving fine details and delivering perceptually pleasing restorations in real-world scenarios.

\section{Ablation Study}
In this section, we use Restormer~\citep{zamir2022restormer} with 16 hidden channels as the baseline model. We maintain the overall architecture, including the total number of Transformer blocks, feed-forward networks (FFNs), and feature fusion strategy, as well as consistent training settings on 4 NVIDIA A100 GPUs. All networks are trained and evaluated on the GoPro dataset~\citep{nah2017deep}, chosen for its ability to ensure stable training across models. We assess restoration quality using both distortion-based metrics (PSNR and SSIM) and perception-based metrics, FID~\citep{heusel2017gans, parmar2021cleanfid}, LPIPS~\citep{zhang2018unreasonable} and NIQE, for comprehensive comparisons. Additionally, we report the total number of parameters and MACs to indicate model complexity.

\setlength{\tabcolsep}{8pt}
\begin{table}[t]
\centering
\caption{Ablation study on dilation factor configurations and the proposed channel-aware self-attention on the GoPro~\citep{nah2017deep} dataset. The baseline is Restormer~\citep{zamir2022restormer} with 16 hidden channels. {\bftab NA} denotes local neighborhood attention while {\bftab DiNA} represents sparse dilated neighborhood attention; with and without are abbreviated as w/ and w/o respectively; {\bftab CAM} is the proposed channel-aware module. The adopted NA-DiNA with CAM method shows the strongest or competitive quantitative visual results as rated by both distortion and perception metrics.}
\begin{tabular}{lcccccc}
\toprule
\multirow{2}{*}{\textbf{Networks}} & \multicolumn{2}{c}{\textbf{Distortion}} & \multicolumn{2}{c}{\textbf{Perception}} & \textbf{Params} & \textbf{MACs} \\
\cmidrule(lr){2-3} \cmidrule(lr){4-5}
 & PSNR ↑ & SSIM ↑ & FID ↓ & LPIPS ↓ & (M) ↓ & (G) ↓ \\
\midrule
\rowcolor{gray!20}
Restormer (baseline)                   
& 30.32 & 0.934 & 15.16 & 0.137 & 3.0 & 17.30 \\
NA w/o CAM                
& 31.56 & 0.948 & 11.12 & 0.111 & 3.0 & \bftab16.64 \\
\rowcolor{gray!20}
NA w/ CAM                        
& 31.97 & 0.952 & 10.86 & 0.105 & 3.0 & 16.71 \\
DiNA w/o CAM  
& 32.03 & 0.953 & 10.17 & 0.103 & 3.0 & \bftab16.64 \\
\rowcolor{gray!20}
DiNA w/ CAM                     
& 32.06 & 0.952 & 10.14 & 0.103 & 3.0 & 16.71 \\
NA-DiNA w/o CAM     
& 31.87 & 0.951 & 10.08 & 0.107 & 3.0 & \bftab16.64 \\
\rowcolor{gray!20}
NA-DiNA w/ CAM                    
& \bftab32.06 & \bftab0.953 & \bftab9.53 & \bftab0.103 & 3.0 & 16.71 \\
\bottomrule
\end{tabular}%
\label{tab:dilation_ablation}
\end{table}

As shown in~\autoref{tab:dilation_ablation}, the local-attention-only network already outperforms the Restormer baseline~\citep{zamir2022restormer} by 1.24 dB in PSNR, despite being the weakest among the proposed configurations. Introducing our channel-aware module further improves the local variant by 0.41 dB, and also enhances the global-only and hybrid variants by 0.02 dB and 0.19 dB, respectively. The hybrid configuration with the channel-aware module achieves the best overall performance across all metrics. These results validate our finding that the original DiNA design~\citep{hassani2022dilated} with hybrid-attention is suboptimal for deblurring, and demonstrate that the proposed channel-aware module effectively addresses this limitation. Moreover, DiNAT-IR retains a similar parameter count while reducing MACs by 0.59G compared to the baseline model. Overall, it achieves a notable 1.74 dB PSNR improvement over the Restormer baseline, offering a superior trade-off between performance and efficiency.

\begin{figure*}[tb]
  \centering
  \includegraphics[width=0.98\linewidth, trim = 0.cm 14.8cm 7.65cm 0.cm]{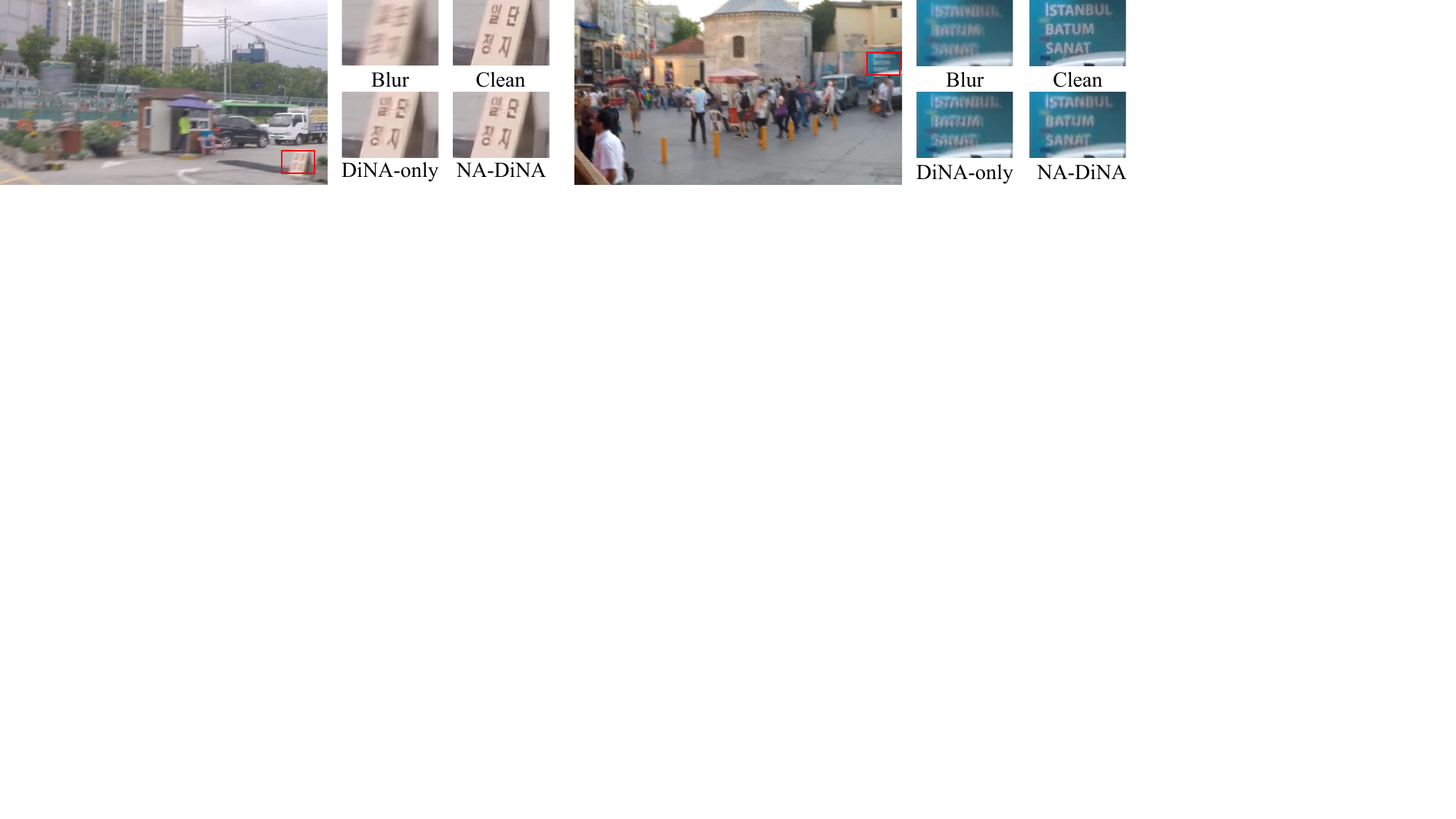}

  \caption{Visual comparisons between DiNAT-IR with global dilated neighborhood attention (DiNA) only and DiNAT-IR with both global DiNA and local neighborhood attention (NA). Both networks are trained and tested on the GoPro dataset~\citep{nah2017deep} with the same training settings.}\label{fig:abl_na}
\end{figure*}

Importantly, although Table~\ref{tab:dilation_ablation} shows that the PSNR, SSIM and LPIPS differences between DiNA with CAM and NA-DiNA with CAM are minimal, our observations reveal that incorporating local neighborhood attention (NA) improves the visual quality of the restored images. As illustrated in Figure~\ref{fig:abl_na}, the method relying solely on global dilated neighborhood attention produces distorted text on the board, whereas including NA results in sharper and more accurate restoration. Therefore, we select DiNAT-IR with NA-DiNA and CAM as our final architecture for the image restoration tasks.

\section{Conclusion}
In this work, we propose DiNAT-IR, a novel Transformer-based architecture for image restoration that effectively balances global context modeling and local detail preservation. Building upon the strengths of Dilated Neighborhood Attention (DiNA), DiNAT-IR introduces a channel-aware module that enhances global context integration while maintaining pixel-level precision of local Neighborhood Attention (NA). Our experiments demonstrate that, although DiNAT-IR does not consistently surpass Restormer~\citep{zamir2022restormer} in all metrics, for instance, achieving slightly lower PSNR on certain deraining benchmarks, it delivers comparable or superior restoration performance, particularly in challenging tasks like motion deblurring and defocus deblurring. Furthermore, DiNAT-IR achieves this high-quality restoration with similar or reduced computational costs, offering a favorable trade-off between restoration quality and efficiency. These findings highlight DiNAT-IR as a promising and versatile solution for diverse low-level vision tasks.
\section{Limitation}
Our ablation studies were conducted primarily on the GoPro dataset~\citep{nah2017deep} to ensure consistent training strategies across all models and enable fair comparison. However, this might limit the generalizability of our results to other datasets. Extending the analysis to broader tasks is non-trivial, as it requires substantial effort to adapt training strategies to different data distributions. Since architectural components may yield varied gains across tasks and dataset bias exists, a universally optimal design remains challenging. Future work will explore task-specific architectures to improve generalization and robustness.
\subsubsection*{Broader Impact Statement}
The proposed DiNAT-IR framework advances the field of image restoration by improving the quality and efficiency of deblurring, deraining, and denoising tasks. Positive societal impacts include potential applications in photography, surveillance, autonomous driving, medical imaging, and digital archiving, where clearer images can enhance safety, usability, and analysis. However, as with many image enhancement techniques, there are potential risks if such technologies are misused to manipulate images, obscure evidence, or produce deceptive visual content. Additionally, improvements in image clarity might inadvertently reveal personal information or sensitive details in images that were previously obscured by poor quality, raising privacy concerns. We encourage future research to consider these ethical implications and to develop safeguards or detection mechanisms to identify manipulated or restored images. Our experiments focus on publicly available datasets, and we do not anticipate direct privacy risks from our work.



\bibliography{tmlr}
\bibliographystyle{tmlr}


\end{document}